\documentclass[runningheads]{llncs}
\usepackage{graphicx}
\usepackage{subfigure}
\usepackage{amsmath}
\usepackage{amssymb}
\usepackage{wrapfig}
\usepackage{upgreek}
\usepackage{hyperref}

\newcommand\blfootnote[1]
{%
\begingroup
\vspace{-5pt}
\renewcommand\thefootnote{}
\footnotetext{#1}%
\endgroup
}

\begin{document}
\title{Spatio-Temporal Graph Convolution for Resting-State fMRI Analysis}
%
%
\author{
Soham Gadgil$^{*,}$\inst{1},
Qingyu Zhao$^{*,}$\inst{2}, 
Adolf Pfefferbaum \inst{2,3}, 
Edith V. Sullivan\inst{2},
Ehsan Adeli\inst{1,2}, 
Kilian M. Pohl\inst{2,3}}
\institute{Computer Science Department, Stanford University, Stanford, USA \and School of Medicine, Stanford University, Stanford, USA \and Center of Health Sciences, SRI International, Menlo Park, USA}

%
\maketitle              
\begin{abstract}
The Blood-Oxygen-Level-Dependent (BOLD) signal of resting-state fMRI (rs-fMRI) records the temporal dynamics of intrinsic functional networks in the brain. However, existing deep learning methods applied to rs-fMRI either neglect the functional dependency between different brain regions in a network or discard the information in the temporal dynamics of brain activity. To overcome those shortcomings, we propose to formulate functional connectivity networks within the context of spatio-temporal graphs. We train a spatio-temporal graph convolutional network (ST-GCN) on short sub-sequences of the BOLD time series to model the non-stationary nature of functional connectivity. Simultaneously, the model learns the importance of graph edges  within ST-GCN to gain insight into the functional connectivities contributing to the prediction. In analyzing the rs-fMRI of the Human Connectome Project (HCP, $N$=1,091) and the National Consortium on Alcohol and Neurodevelopment in Adolescence (NCANDA, $N$=773), ST-GCN is significantly more accurate than common approaches in predicting gender and age based on BOLD signals. Furthermore, the brain regions and functional connections significantly contributing to the predictions of our model are important markers according to the neuroscience literature.  

\end{abstract}
\section{Introduction}
\blfootnote{* Equal contribution}
The BOLD signal of rs-fMRI characterizes the intrinsic functional organization of the human brain by measuring its spontaneous activity at rest \cite{Buckner13}. One commonly used approach for identifying impact of factors, such as age and sex, on intrinsic functional networks is to predict their values by applying deep neural networks to the rs-fMRI of individual subjects   \cite{Ktena17,Li19,Dvornek17,Li18}. Accurate predictors of those factors can then enhance the understanding of functional neurodevelopment across the life span \cite{Li18b}, characterize developmental disruption caused by neurological disorders, and explain sex-specific differences in cognitive performance\cite{Weis19}. 

It is not trivial to select an appropriate network architecture for analyzing rs-fMRI as their (average) BOLD signals of each brain region are structured time series. A natural representation of such data are spatio-temporal graphs \cite{Yan18,Yu18}, where the temporal graph characterizes the dynamics of brain activity at each region and the spatial graph characterizes the functional interaction between different brain regions. However, most existing deep learning works applied to rs-fMRI analysis fail to consider both aspects simultaneously. Methods that only incorporate spatial graph convolution \cite{Ktena17,Li19,Zhang19} often transform the time series data into hand-crafted features, such as partial correlation \cite{marrelec2006partial}, thereby potentially losing the fine temporal information of the BOLD signal. Methods based on recurrent neural networks (RNN) \cite{Dvornek17,Li18,Cui18} can learn temporal features from the BOLD signal but neglect the functional dependency between regions-of-interests (ROIs). On the other hand, methods based on spatio-temporal networks often perform spatial convolution according to the topological arrangement among ROIs in the physical space \cite{Ktena17,Zhao2018stcnn}, which cannot model interactions among distal ROIs with similar functional properties. To address these issues, we suggest deep neural networks to incorporate spatio-temporal convolution on functional connectivity graphs, i.e., spatio-temporal Graph Convolution (ST-GC) \cite{Yan18,Yu18}.

In computer vision, ST-GC Networks (ST-GCN) are popular for solving problems that base prediction on graph-structured time series \cite{Yan18,Yu18,Covert19}. In the context of rs-fMRI analysis, these networks have the potential to automatically extract features that jointly characterize functional connectivity patterns of the brain and their temporal dynamics within the BOLD series. To the best of our knowledge, we are the first to use ST-GCN for building predictive models based on rs-fMRI data. We train our network on short sub-sequences of BOLD time series to model the non-stationary nature of functional connectivity \cite{Taghia17,Zhao19}. Further, learning of an edge importance matrix associated with the ST-GC operation improves the interpretability of the model as it allows us to identify selective functional connections significantly contributing to the prediction. We apply ST-GCN to predict the age and gender of healthy individuals of two large publicly available rs-fMRI datasets: Human Connectome Project (HCP, $N$=1,096) \cite{van2013wu} and National Consortium on Alcohol and Neurodevelopment in Adolescence (NCANDA, $N$=773) \cite{Brown2015}. The optimal window size for the sub-sequences highly coincide between the two datasets despite their distinct imaging protocols and data processing pipelines. Furthermore, the resulting prediction accuracy is significantly higher than traditional RNN-based methods. Finally, the learned edge importance localizes meaningful brain regions and functional connections associated with aging effects and sexual differences.\footnote{The code is available at \url{https://github.com/sgadgil6/cnslab_fmri}}

\section{ST-GCN for rs-fMRI Analysis} 
We first relate functional networks to spatio-temporal graphs and then define the ST-GC convolution on them. Next, we build the classifier ST-GCN, which uses ST-GC in each layer and models the importance of graph edges in the decision process across layers. Lastly, we train ST-GCN on short BOLD sequences.

\begin{figure}[t]
\centering
\includegraphics[width=0.8\textwidth]{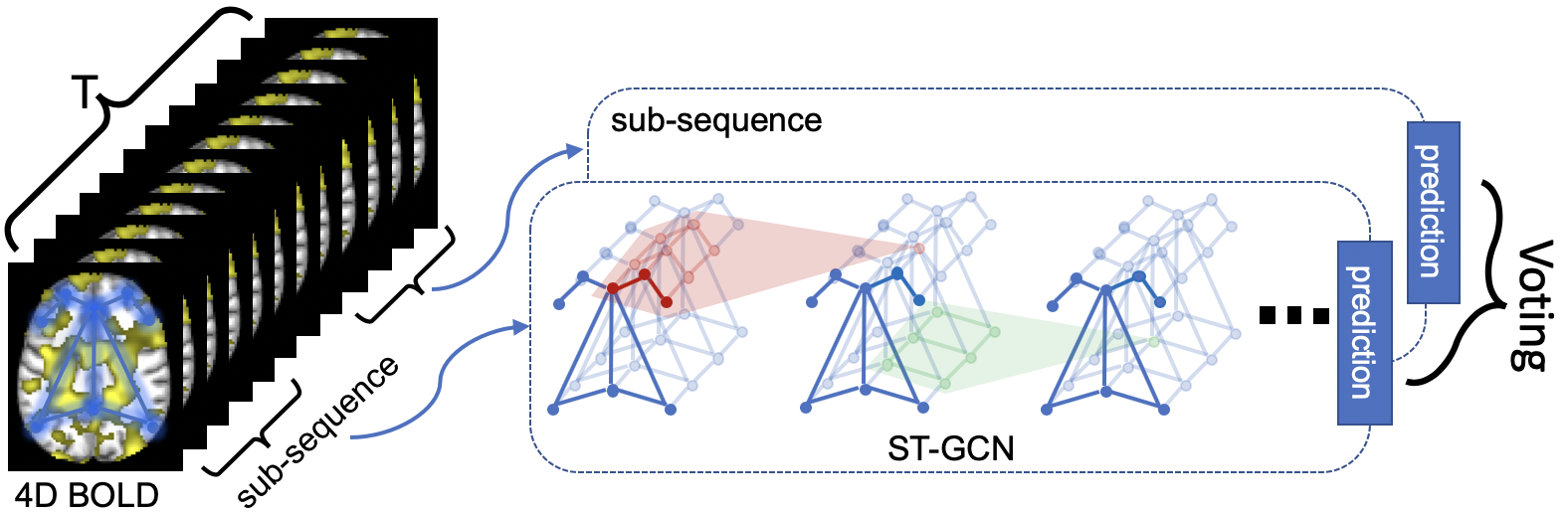}
\caption{Framework of classifying BOLD time series by applying a spatio-temporal graph convolutional network to sub-sequences.} \label{fig:fig1}
\end{figure}

\paragraph{\bf Representing Functional Networks as Spatio-Temporal Graphs:}  To encode the functional networks captured by rs-fMRI, let $\mathcal{G}:=(\mathcal{V},\mathcal{E})$ (Fig. \ref{fig:fig1}) be an undirected spatio-temporal graph consisting of a set of edges $\mathcal{E}$ capturing temporal and spatial connections between a set of nodes $\mathcal{V}=\{v_{t,i}|t=1,...,T; i=1,...,N\}$ defined across $N$ ROIs and $T$ time points. For each ROI and time point, an edge in the {temporal graph}  connects the corresponding node to the node of the same ROI at the proceeding time point. The edges of the {spatial graph} connect all nodes of the same time point, where the weight of an edge is defined by the functional affinity between the corresponding regions. To define the functional affinity, we now assume that spontaneous activation for each of the $N$ regions can be quantified by the average BOLD time series measured within that region. First,  those series are concatenated across all subjects within each ROI. Then, the affinity between two regions $d(v_{tj}, v_{ti})$ is defined by the magnitude of correlation between their concatenated series. Note, this affinity is impartial to the time point.

\paragraph{\bf Spatio-Temporal Graph Convolution (ST-GC):} To define a convolution on such a  graph structure, we denote $f_{in}(v_{ti})$ as the input feature at node $v_{ti}$ (e.g., the average BOLD signal of ROI $i$ at time $t$) and $B(v_{ti})$ as the spatio-temporal neighbourhood of $v_{ti}$, i.e., 
\begin{equation}
    B(v_{ti}):= \{ v_{qj} | d(v_{tj}, v_{ti}) \leq K, |q-t| \leq \lfloor \Gamma / 2 \rfloor \},
\label{eq:neighbourhood}
\end{equation}
where $K$ defines the size of the spatial neighborhood (i.e., spatial kernel size) and $\Gamma$ the temporal neighborhood (i.e., temporal kernel size). An ST-GC operation on node $v_{ti}$ with respect to a convolutional kernel $\textbf{w}(\cdot)$ and a normalization factor $Z_{ti}$ can then be defined as \cite{Yan18}
\begin{equation}
    f_{out}(v_{ti}):=\frac{1}{Z_{ti}} \sum_{v_{qj} \in B(v_{ti})}  f_{in}(v_{qj})\cdot \textbf{w}(v_{qj}).
\label{eq:stgc}
\end{equation}

Adopting a similar implementation as in \cite{Yan18}, we approximate the spatio-temporal convolutional kernel $\textbf{w}(\cdot)$ by decomposing it to a spatial graph convolutional kernel $\textbf{W}_{SG} \in \mathbb{R}^{C \times M}$ represented in the Fourier domain and a temporal convolutional kernel $\textbf{W}_{TG} \in \mathbb{R}^{M \times \Gamma}$. Specifically, we denote $\textbf{f}_t \in \mathbb{R}^{N \times C}$ as the $C$-channel input features of the $N$ ROIs at the $t^{th}$ frame, $\textbf{f}_{t}' \in \mathbb{R}^{N \times M} $ as the $M$-channel output features, and $\textbf{A}$ as the aforementioned affinity matrix. The spatial graph convolution at a time point $t$ is then defined with respect to the diagonal matrix $\boldsymbol{\Lambda}$ (where $\Lambda^{ii}=\sum_j A^{ij}+1$) as \cite{Kipf17}
\begin{equation}
    \textbf{f}_t' := \boldsymbol{\Lambda}^{-\frac{1}{2}}(\textbf{A}+\textbf{I})  \boldsymbol{\Lambda}^{-\frac{1}{2}}\textbf{f}_t \textbf{W}_{SG}.
\label{eq:sgc}
\end{equation}
Next, the temporal convolution is performed on the resulting features. Let $\textbf{f}'_i \in \mathbb{R}^{M \times T}$ be the features of node $v_i$ defined on the temporal graph of length $T$ with regular grid spacing. We then perform a standard 1D convolution $\textbf{f}'_i \circledast \textbf{W}_{TG} \in \mathbb{R}^{\Gamma \times T}$ as the final output of ST-GC for $v_i$.

\paragraph{\bf Classifying BOLD Time Series by ST-GCN:} Our proposed ST-GCN is composed of 3 layers of ST-GC units. The input to ST-GCN are 1-channel spatio-temporal features $\textbf{f}\in \mathbb{R}^{N \times T}$ representing the average BOLD signals of the $N$ ROIs. Consistent with the setup in \cite{Yan18}, each ST-GC layer produces 64-channel outputs with the temporal kernel size $\Gamma=11$, a stride of 1, and a dropout rate of 0.5. The output of the last ST-GC layer is fed to a global average pooling and its output vector of length 64 is transformed to {\it class} probabilities by a fully connected layer with a sigmoid activation. 

To determine the importance of spatial graph edges in defining class probabilities, we integrate a positive and symmetric ``edge importance" matrix $\textbf{M} \in \mathbb{R}^{N \times N}$ into our model. This matrix is shared across all ST-GC layers by replacing $\textbf{A}+\textbf{I}$ in Eq. \eqref{eq:sgc} with $(\textbf{A}+\textbf{I})*\textbf{M}$ (element-wise multiplication). As such, while performing spatial graph convolution on node $v_{ti}$, the contribution from its neighbouring nodes $B(v_{ti})$ will be re-scaled according to the importance weights learned in the $i^{th}$ row of $\textbf{M}$. Thus, the diagonal entries of $\textbf{M}$ (self-connection) quantify importance for each ROI, while off-diagonal entries do so for each functional connection. Note, the original proposal by Yan et al. \cite{Yan18} learns a separate $\textbf{M}$ for each individual ST-GC layer. Their strategy generally  results in negative, asymmetric importance matrices that are difficult to interpret and vary across layers. We ease the interpretability by enforcing $\textbf{M}$ to be consistent across layers and be both positive and symmetric.

\paragraph{\bf Training ST-GCN:} Recent rs-fMRI studies have revealed that patterns of intrinsic functional connectivity are not stationary across the full rs-fMRI scan but exhibit considerable fluctuations. These dynamics are often analyzed by dividing the entire rs-fMRI sequence into sub-sequences according to a fixed window size (often chosen empirically) and then assessing the connectivity within these segments. Accordingly, we also consider training ST-GCN on sub-sequences of window size $T'$ sampled from the full data. Specifically, at each training iteration, we sample a sub-sequence of length $T'$ starting at a random time frame from the full sequence of each training subject in the mini-batch. The models are then trained by stochastic gradient descent with a learning and weight decay rate of 0.001. At the testing stage, we sample $S$ sub-sequences, each starting at a random time frame, of length $T'$ for each testing subject, derive the ST-GCN prediction for each sub-sequence, and perform a simple voting to produce the final subject-level prediction, e.g., the average of sigmoid values in the case of binary classification (Fig. \ref{fig:fig1}).

\section{Experiments}
Understanding the dramatic neurodevelopment and the emerging sexual differences during adolescence is an important topic in neuroscience \cite{Weis19,Smith13}. We investigate the age and sex difference using ST-GCN and a variety of baseline approaches on the NCANDA dataset. We further investigate sex differences in young adults of the Human Connectome Project (HCP) S1200 \cite{van2013wu}. Note, identifying significant aging effects on HCP is unlikely due to functional organization reaching maturity after young adulthood \cite{Westlye10}.

\paragraph{\bf NCANDA:} The publicly released baseline data consisted of 773 rs-fMRI scans (269 frames, TR=2s). Among the 773 adolescents (ages 12-21 years, 388 younger than the the mean age of 16 years vs. 385 older adolescents, 376 male vs. 397 females). 638 met the no-to-low alcohol drinking criteria of NIAAA (373 young vs. 265 old adolescents, 315 boys vs. 323 girls) \cite{Brown2015}. Each rs-fMRI scan was preprocessed by the NCANDA pipeline, which registered the mean BOLD image to subject-specific T1 MRI and then non-rigidly to the standard MNI space \cite{Zhao19b}. The cortical surface was  parcellated to $N=34$ ROIs according to \cite{Klein2012}. The average BOLD signal in each ROI was normalized to \textit{z}-scores.

\paragraph{\bf HCP:} The data set consisted of  rs-fMRI of 1096 young adults (ages 22-35 years). Excluding 5 rs-fMRIs with less than 1200 frames, we used the first session (15 min, $T=1200$ frames, TR=0.72s) for each of the  498 females and 593 males. Each rs-fMRI went through the minimal processing pipeline of HCP, \textit{fMRISurface} \cite{Glasser13}, which mapped each volume time series to the standard CIFTI grayordinates space. The cortical surface was parcellated to $N=22$ major ROIs \cite{Glasser16}, and the average BOLD signal in each ROI was normalized to \textit{z}-scores. 

\paragraph{\bf Experimental Setup:}
On NCANDA, ST-GCN first distinguished younger from older participants by performing 5-fold cross-validation on the 773 rs-fMRIs. The training and testing were repeated on sub-sequences of different window sizes, from short segments of $T'=16$ to the full sequence. The number of sub-sequences used for voting in the testing stage was fixed at $S=64$ ($S=1$ for the full sequence). To ensure the prediction results were not confounded by alcohol drinking, the test accuracy was also recorded for the no-to-low drinking cohort, denoted as ``ST-GCN-no-ex". Based on the optimal window size determined through cross-validation, we trained the model on the entire data set to produce an edge importance matrix summarizing the aging effects within the entire cohort. To reduce uncertainty in the estimation caused by stochastic gradient descent, we repeated the training 20 times and derived the ``average" edge importance matrix as our final outcome. Next, this experiment  was repeated with respect to sex classification on NCANDA and HCP. 

\paragraph{\bf Baselines:} The simplest approach for our comparison was a Multi-Layer Perceptron ({\bf MLP}) applied to the upper triangular correlation matrix, which quantified the static functional connectivity between ROIs during the full scan. The MLP had 2 hidden layers, each with 64 neurons and ReLu activation. The next two baseline methods relied on end-to-end training on the BOLD time series based on  Long Short-Term Memory (LSTM) model \cite{hochreiter1997long}, a form of RNN frequently used to analyze rs-fMRI data \cite{Dvornek17,Li18}. Adopting a similar implementation as in \cite{Li18}, the first implementation of {\bf LSTM} consisted of a recurrent cell with hidden states of dimension 256. The output of the last hidden state was fed into a fully connected layer (with a dropout of 0.5) to produce a final label. The second implementation ({\bf GC-LSTM}) first extracted features from the BOLD signal at each time frame using spatial graph convolution. The features were then fed into an LSTM model for temporal analysis. The number of voting sequences for these two LSTM-based methods were again fixed to $S$=64. 

\section{Results and Analysis}
\begin{figure}[t]
\includegraphics[width=\textwidth]{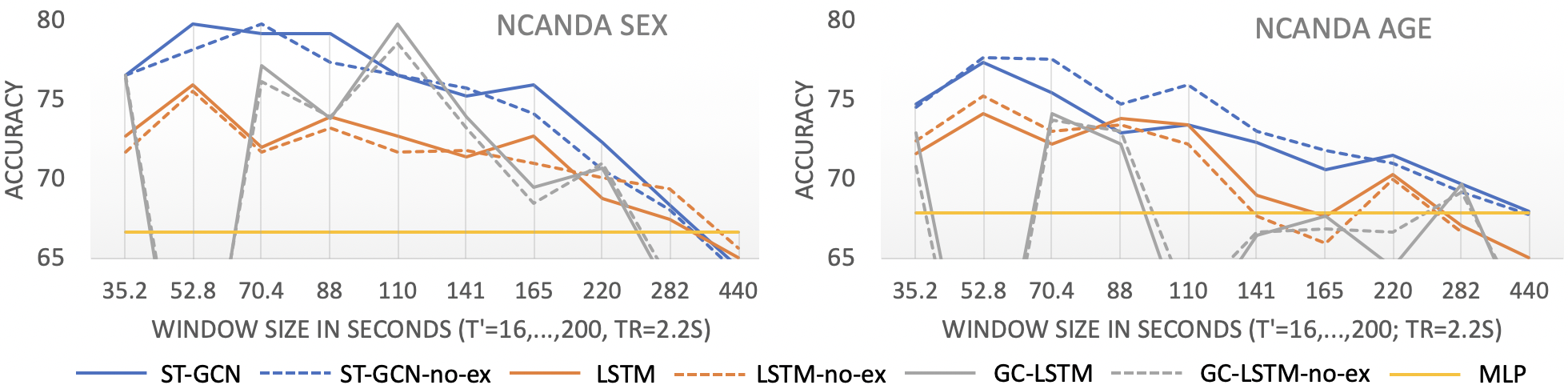}
\caption{Accuracy of age and sex prediction on the NCANDA dataset. ``no-ex" denotes the accuracy score confined to the no-to-low drinking cohort.} \label{fig:ncanda_acc}
\end{figure}

\paragraph{\bf NCANDA:} Fig. \ref{fig:ncanda_acc} shows the prediction accuracy derived from the 5-fold cross-validation for different methods. The accuracy scores measured on all NCANDA subjects were not significantly different from results on the no-to-low drinking cohort (denoted as ``no-ex", paired \textit{t}-test p$>$0.5), indicating that the predictions were not confounded by alcohol consumption in individuals. On both data sets,  all three deep-learning-based methods were generally more accurate than MLP when applied to shorter sub-sequences. ST-GCN achieved significantly higher accuracy than the two LSTM-based methods across different window sizes (p$<$.0001, two-sample \textit{t}-test). Performing further architecture search for the two LSTM-based methods, such as varying the hidden-state dimension and adding more LSTM layers, did not improve the results in Fig. \ref{fig:ncanda_acc}. Moreover, adding more convolution layers or feature channels to GC-LSTM reduced the accuracy substantially, which indicated that the strategy of training spatial graph convolution on each individual frame without incorporating temporal convolution was flawed. With respect to the no-to-low drinking participants, the highest prediction accuracy of ST-GCN with respect to age was  77.7\% ($T'=24$ or 52.8 s) and sex was 79.8\% ($T'=32$ or 70.4 s). The accuracy of sex prediction was relatively stable across $T'=[24,40]$ (52.8 - 88 s). These predictions were not confounded by subject motion based on the insignificant correlation between the prediction scores and the number of outlier frames in each rs-fMRI.

\begin{figure}[t]
\centering
\includegraphics[width=\textwidth]{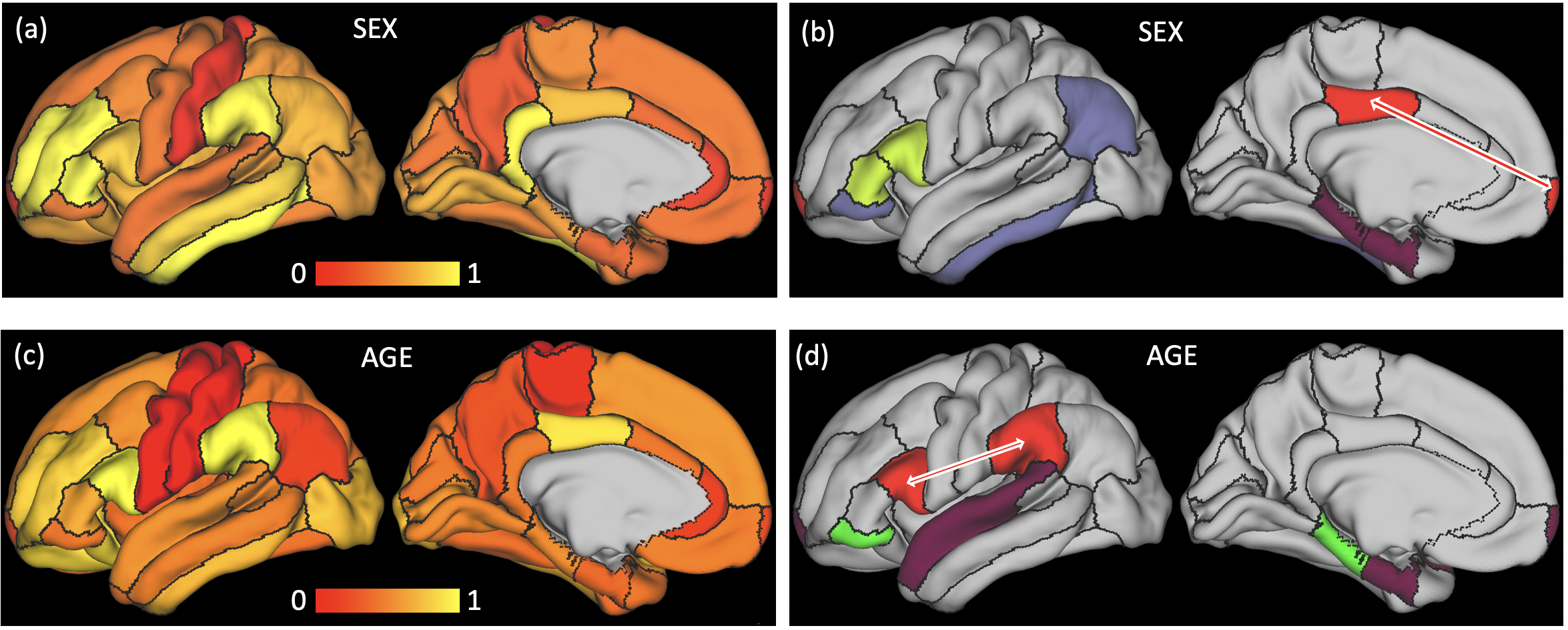}
\caption{Importance maps for age (a,b) and sex (c,d) prediction of the NCANDA study. (a,c): Importance of functional dynamics within each ROI; (b,d): Functional connections between ROIs with importance higher than 0.3 are shown by the ROIs having the same color. Highlighted by the red arrows are the default mode network for sex prediction and the inferior temporo-parieto-frontal network for age prediction.} \label{fig:ncanda_sex_age}
\end{figure}

We visualized the edge importance matrices associated with the optimal models in Fig. \ref{fig:ncanda_sex_age} with the images on the left (Fig. \ref{fig:ncanda_sex_age}a+c) showing the importance of each ROI (diagonal entries) for prediction and the images on the right (Fig. \ref{fig:ncanda_sex_age}b+d) showing functional connections (off-diagonal entries) with importance value higher than 0.3. For sex prediction, the most important ROI identified by ST-GCN was the inferior temporal lobe, which echoed findings from other resting-state studies \cite{Conrin18} and from a structural MRI analysis on the NCANDA cohort \cite{Zhao19c}. We also identified a significant effect in the frontal-posterior-cingulate (PCC) connection (red in Fig.  \ref{fig:ncanda_sex_age}b), which defines the default mode network, a signature intrinsic network frequently linked to sexual differences \cite{MO17}. For age prediction, the most critical ROIs were the supramarginal and par opercularis regions (Fig. \ref{fig:ncanda_sex_age}c). Their functional connection (red in Fig. \ref{fig:ncanda_sex_age}d) defined the inferior temporo-parieto-frontal network, which was shown to decrease in older adolescents within the NCANDA cohort by a longitudinal rs-fMRI study \cite{Zhao19b}. All the results above demonstrate that our strategy of edge importance learning can accurately localize functional properties of the brain related to significant aging and sex effects.

\begin{figure}[t]
\includegraphics[width=\textwidth]{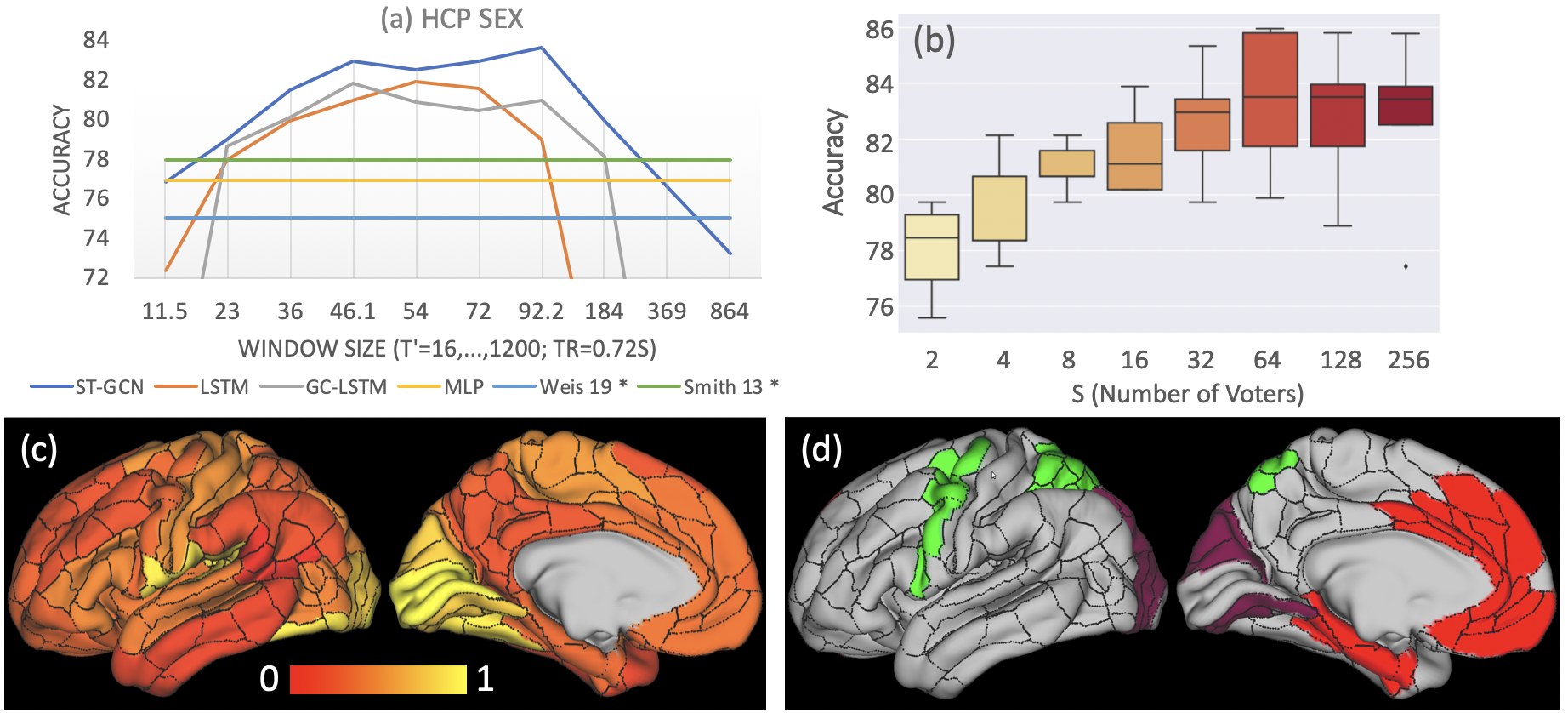}
\caption{Experiments on HCP: (a) Sex prediction accuracy w.r.t window size. * denotes results from prior studies. (b) Distribution of prediction accuracy scores from cross-validation w.r.t. the number of sub-sequences used for voting. (c) Importance of functional dynamics within each ROI; (d) Functional connections between ROIs with importance higher than 0.3 are shown by the ROIs having the same color.} \label{fig:hcp_results}
\end{figure}

\paragraph{\bf HCP:} Fig. \ref{fig:hcp_results}a plots the accuracy in predicting sex on the HCP dataset in relation to those of prior studies by Weis et al. \cite{Weis19} and Smith et al. \cite{Smith13}. These works applied linear classifiers and used correlation coefficients as input features respectively, which was similar to MLP. Note that their results can not be strictly compared to ours as different cohorts and data processing pipelines were used for analysis \footnote{the 87\% accuracy in \cite{Smith13} on 131 HCP subjects was based on multi-modal data}. However, their  accuracies were similar to the ones of MLP, which were suboptimal compared to those produced by the deep-learning-based methods applied to sub-sequences of $T' \in [32,128]$ (23.04 - 92.16 s). This finding indicates that the dynamical properties of functional interactions among brain regions can not be fully captured by the static correlation coefficients and require more comprehensive spatio-temporal modeling.

An example of more comprehensive spatio-temporal modeling was ST-GCN, which achieved the highest accuracy of 83.7\% at $T'=128$ (92.16 s) and produced higher accuracies than the two LSTM-based methods across all window sizes. Increasing the number of voters did not further increase testing accuracy (Fig. \ref{fig:hcp_results}b). Moreover, ST-GCN produced similarly optimal accuracy scores for $T' \in [64,128]$ (46.08 - 92.16 s). This range was highly consistent with the one revealed in the NCANDA experiment despite the difference in imaging protocols, length of BOLD signal, brain parcellation, and data processing pipelines between the two studies. These results also converge with recent understanding in neuroscience literatures stating that dynamical functional connectivity generally have dwell times of tens of seconds \cite{Allen12,Taghia17}. By visualizing the edge importance matrix derived from the HCP subjects (Fig. \ref{fig:hcp_results}c), we found that regions with significant sexual differences in young adults were spatially more concentrated compared to adolescents (the NCANDA experiment) and mainly located in the visual cortex. This phenomenon could be potentially linked to the evidence that sexual dimorphism in intrinsic functional organization diminishes with age (sex-age interaction) during adolescence \cite{MO17}.

\section{Conclusion}
We introduced a framework for analyzing rs-fMRI data based on spatio-temporal graph convolution networks. By analyzing the rs-fMRI data of two large-scale neuroimaging studies, we showed that ST-GCN could accurately predict age and gender of the study participants based on short sequences of BOLD time series. The similar optimal window sizes (of the sub-sequences) between the two datasets highlighted the usage of short BOLD sequences for modeling dynamic functional connectivity. Future work will focus on defining ST-GC with respect to non-static graph structures accommodating dynamic functional states, exploring automatic determination of sliding window size, applying ST-GCN to a fine-grained brain parcellation, and identifying functional biomarkers linked to neuropsychiatric disorders. Accomplishing these goals will then show if our strategy for learning graph edge importance within the context of model interpretation is valuable for advancing knowledge in neuroscience. 

\textbf{Acknowledgment}. This research was supported in part by NIH grants AA021697, AA005965, AA013521, and AA010723.

%
%
%
\bibliographystyle{splncs04}
{ \small
\bibliography{ref}
}
\end{document}